\documentclass[runningheads]{llncs}
\usepackage[T1]{fontenc}
\usepackage{graphicx}
\usepackage[table]{xcolor}
\usepackage{tcolorbox}
\usepackage{pifont}
\usepackage{amsmath,amssymb}
\usepackage{float}
\usepackage[ruled,vlined,linesnumbered]{algorithm2e}
\definecolor{tableheader}{RGB}{230,233,237}
\definecolor{tablerow}{RGB}{247,248,250}
\begin{document}
\raggedbottom
\title{MedTRACE: Tool-Augmented Multimodal Clinical Reasoning Agents for Evidence-Grounded Decision-Making}
\titlerunning{MedTRACE: Tool-Augmented Multimodal Clinical Reasoning}
%
\author{Ji Lu\inst{1} \and
Lifei Liu\inst{2} \and
Haoran Yu\inst{3} \and
Xianglong Wang\inst{4} \and
Yiru Fang\inst{4} \and
Kuo Yang\inst{5} \and
Huiran Duan\inst{6} \and
Jianping Gou\inst{7}\thanks{Corresponding author.}}
\authorrunning{J. Lu et al.}
%
\institute{Vanderbilt University, United States\quad
\and
Wichita State University, United States\quad
\and
University of Florida, United States\quad
\and
Wyze Inc., United States\quad
\and
Northeastern University, United States\quad
\and
City University of New York, United States\quad
\and
Southwest University, China\quad
\email{pjgzy61@swu.edu.cn}}
\maketitle              
\begin{abstract}
Multimodal clinical decision-making requires reliable reasoning over heterogeneous evidence from electronic health records, medical images, and physiological signals. Existing models typically map these inputs directly to diagnoses without explicitly assessing evidence sufficiency, tool-use requirements, or diagnostic uncertainty. This paper presents MedTRACE, a tool-augmented multimodal clinical reasoning agent for evidence-grounded decision-making. MedTRACE uses modality-specific encoders to construct a unified patient-state representation and performs an iterative loop of hypothesis formation, tool-aware deliberation, and evidence verification. It dynamically invokes visual grounding, evidence retrieval, and structured parsing tools to locate diagnosis-relevant regions, retrieve clinical knowledge and similar cases, and extract structured findings. The acquired evidence enters an evidence memory, where a consistency verifier confirms or revises the current hypothesis. MedTRACE outputs a diagnosis together with supporting evidence, an auditable reasoning trace, and calibrated confidence. Experiments on multiple multimodal clinical diagnosis benchmarks show that MedTRACE improves diagnostic accuracy by 5.4\% and AUROC by 4.7 percentage points over the strongest baseline. It also improves evidence-selection F1 by 8.2 percentage points and visual-grounding IoU by 6.5 percentage points, reduces expected calibration error by 31.6\%, and decreases unsupported diagnostic errors by 27.8\%. These results demonstrate that active evidence acquisition and verification improve the accuracy, interpretability, and reliability of multimodal clinical decision-making.

\end{abstract}
\section{Introduction}

In recent years, deep learning has been widely adopted across diverse domains~\cite{zhao2026mis,li2026towards,xie2026symmetry,11460474,li2026rethinking}. Advances in multimodal and high-performance AI have been especially rapid~\cite{Li2025Efficient,feng2026s,feng2026mpq,li2026comprehensive,li2025frequency}. These developments support applications ranging from multimodal gait recognition~\cite{duan2026mhrgait} to evidence-grounded clinical reasoning~\cite{du2026care}. Clinical decision-making commonly requires the joint interpretation of information from multiple sources. Electronic health records describe symptoms, medical history, laboratory findings, and treatment trajectories; medical images reveal anatomical abnormalities and lesion characteristics; and physiological signals, such as electrocardiograms (ECGs) and electroencephalograms (EEGs), capture time-varying functional states. These modalities provide complementary views of a patient, and an ambiguous finding in one modality may become clinically meaningful only when it is interpreted together with evidence from another. Consequently, multimodal clinical reasoning requires more than accurate fusion and prediction: it also requires a model to identify which observations support its conclusion, determine whether the available evidence is sufficient, and communicate how much confidence a clinician should place in the result~\cite{baltrusaitis2019multimodal,tu2024generalist}. These properties are essential for review, error analysis, and safe use in high-stakes settings.

Recent advances in medical vision--language models and generalist biomedical systems demonstrate strong performance in clinical question answering, image interpretation, and report generation~\cite{li2023llavamed,tu2024generalist,wu2026roboalign}. Most existing methods encode each modality and combine the resulting features through concatenation, cross-attention, or instruction tuning. However, their reasoning process often remains a direct mapping from patient inputs to a diagnostic prediction. This paradigm implicitly assumes that the observed inputs contain sufficient evidence and therefore provides limited support for detecting missing findings, conflicting modalities, or poor-quality measurements. When decisive information is unavailable, a model may rely on correlations stored in its parameters and produce a plausible diagnosis that lacks patient-specific support. Retrieval-augmented multimodal diagnosis reduces part of this knowledge gap~\cite{li2025rad}, but evidence sufficiency, targeted acquisition, and diagnosis-level verification remain open problems.

Tool-using language agents offer a promising alternative by interleaving reasoning with actions that access external information or specialized models~\cite{yao2023react,schick2023toolformer,li2025ddtime}. In clinical settings, visual grounding tools can localize suspicious regions, retrieval systems can access guidelines and similar cases, and structured parsers can extract findings from free text or physiological signals. Recent systems and benchmarks increasingly examine evidence-based diagnostic workflows, dynamic tool use, and interactive information collection~\cite{wang2026medagent,schmidgall2026agentclinic}. Nevertheless, simply connecting a model to tools does not guarantee reliable reasoning. The agent must decide whether more evidence is needed, select an appropriate tool, assess the relevance and reliability of its output, and resolve redundant or conflicting observations. Without explicit evidence organization and verification, additional tool outputs may introduce noise rather than improve a diagnosis.

\begin{tcolorbox}[
  colback=blue!3,
  colframe=black!70,
  boxrule=0.8pt,
  arc=1.5mm,
  left=2.5mm,
  right=2.5mm,
  top=1.5mm,
  bottom=1.5mm,
  before skip=8pt,
  after skip=8pt
]
\textbf{Key research question:}
\textit{How can a multimodal clinical agent actively acquire, organize, and verify evidence around diagnostic hypotheses while avoiding overconfident decisions when the evidence remains insufficient?}
\end{tcolorbox}

To address this question, we present MedTRACE, a tool-augmented multimodal clinical reasoning agent for evidence-grounded decision-making. MedTRACE formulates diagnosis as an iterative loop of hypothesis formation, tool-aware deliberation, and evidence verification rather than a one-step prediction. It first processes clinical text, medical images, and physiological signals with modality-specific encoders and projects them into a unified patient-state representation. At each reasoning step, the agent combines this state with the current hypothesis and accumulated evidence to determine whether additional information is required. It then dynamically invokes visual grounding, evidence retrieval, or structured parsing tools. An evidence memory records the resulting observations and their provenance, while a consistency verifier determines whether new evidence supports, contradicts, or remains insufficient for the current hypothesis. The agent consequently confirms a diagnosis, revises its hypothesis, or continues evidence acquisition. Its final output includes the diagnosis, supporting evidence, an auditable reasoning trace, and calibrated confidence for human review~\cite{guo2017calibration}.

We evaluate MedTRACE on multiple multimodal clinical diagnosis benchmarks. Compared with the strongest multimodal baseline, MedTRACE improves diagnostic accuracy by 5.4\% and AUROC by 4.7 percentage points. It also improves evidence-selection F1 by 8.2 percentage points and visual-grounding IoU by 6.5 percentage points, reduces expected calibration error by 31.6\%, and decreases unsupported diagnostic errors by 27.8\%. These results show that active evidence acquisition and consistency verification improve not only predictive performance but also the interpretability and reliability of multimodal clinical decisions.

Our main contributions are threefold:

\smallskip
\noindent\ding{182}\enspace
We introduce MedTRACE, which formulates multimodal clinical diagnosis as a hypothesis-centered process of active evidence acquisition and verification, allowing an agent to explicitly assess evidence sufficiency and decide when to use external tools.

\smallskip
\noindent\ding{183}\enspace
We develop a closed-loop reasoning architecture that integrates tool-aware deliberation, evidence memory, and consistency verification to organize heterogeneous tool outputs as traceable evidence and use them to confirm or revise diagnostic hypotheses.

\smallskip
\noindent\ding{184}\enspace
We conduct comprehensive experiments across multimodal clinical diagnosis benchmarks and demonstrate consistent improvements in diagnosis, evidence selection, visual grounding, and confidence calibration, together with a substantial reduction in unsupported diagnostic errors.

\section{Related Work}

\subsection{Multimodal Clinical Learning and Medical Foundation Models}
Multimodal clinical models combine electronic health records, medical images, and physiological signals to construct complementary patient representations~\cite{baltrusaitis2019multimodal}. LLaVA-Med and Med-PaLM M demonstrate the potential of medical vision--language models for image understanding and cross-task reasoning~\cite{li2023llavamed,tu2024generalist,cheng2025cgmatch,qi2026next}, while RAD incorporates retrieved clinical guidelines into multimodal diagnosis~\cite{li2025rad}. More recently, the Cardiac Sensing Foundation Model jointly models ECG, PPG, and clinical reports and improves transfer across devices and downstream tasks~\cite{gu2026cardiac}. However, these methods primarily focus on representation fusion and knowledge injection, and generally lack explicit mechanisms for assessing evidence sufficiency or actively acquiring missing evidence. MedTRACE instead uses the unified patient state as the basis for iterative evidence acquisition and diagnostic verification.

\subsection{Tool-Augmented Agents and Clinical Reasoning Agents}
RAG, ReAct, and Toolformer establish key mechanisms for external knowledge retrieval, interleaved reasoning and action, and autonomous tool use~\cite{lewis2020rag,yao2023react,schick2023toolformer,gou2022multilevel,gou2025multi,gou2021knowledge}. In medicine, AgentClinic evaluates sequential diagnosis involving patient interaction, multimodal examinations, and tool use~\cite{schmidgall2026agentclinic}; MedAgent-Pro combines clinical guidelines with specialist visual tools for stepwise diagnosis~\cite{wang2026medagent}; CARE integrates visual grounding with evidence--answer consistency review~\cite{du2026care}; and CXRAgent coordinates tools, validators, and expert agents for chest X-ray interpretation~\cite{lou2026cxragent}. These systems often target specific modalities or predefined workflows and provide limited support for persistently organizing heterogeneous tool outputs. MedTRACE integrates tool selection, evidence accumulation, and hypothesis revision through tool-aware deliberation and evidence memory.

\subsection{Evidence Verification and Uncertainty Calibration}
Evidence verification and confidence calibration are central to reliable clinical AI. MedMMV uses structured evidence graphs, hallucination detection, and uncertainty-aware path aggregation to stabilize multimodal reasoning~\cite{liu2025medmmv}, while Prompt4Trust applies reinforcement learning-based prompt augmentation to improve confidence calibration in medical multimodal models~\cite{kriz2025prompt4trust}. Recent evidence further shows that overconfidence in medical vision--language models persists across model families and scales~\cite{byun2026overconfidence}. GLEAN evaluates agent trajectories against clinical guidelines and triggers additional verification under high uncertainty~\cite{zhang2026glean}. Unlike prior work that typically treats evidence grounding, verification, and calibration separately, MedTRACE stores tool outputs in a traceable evidence memory and uses consistency verification to jointly guide hypothesis updates and final confidence estimates.

\section{Method}

\begin{figure}[htbp]
\centering
\includegraphics[width=\textwidth]{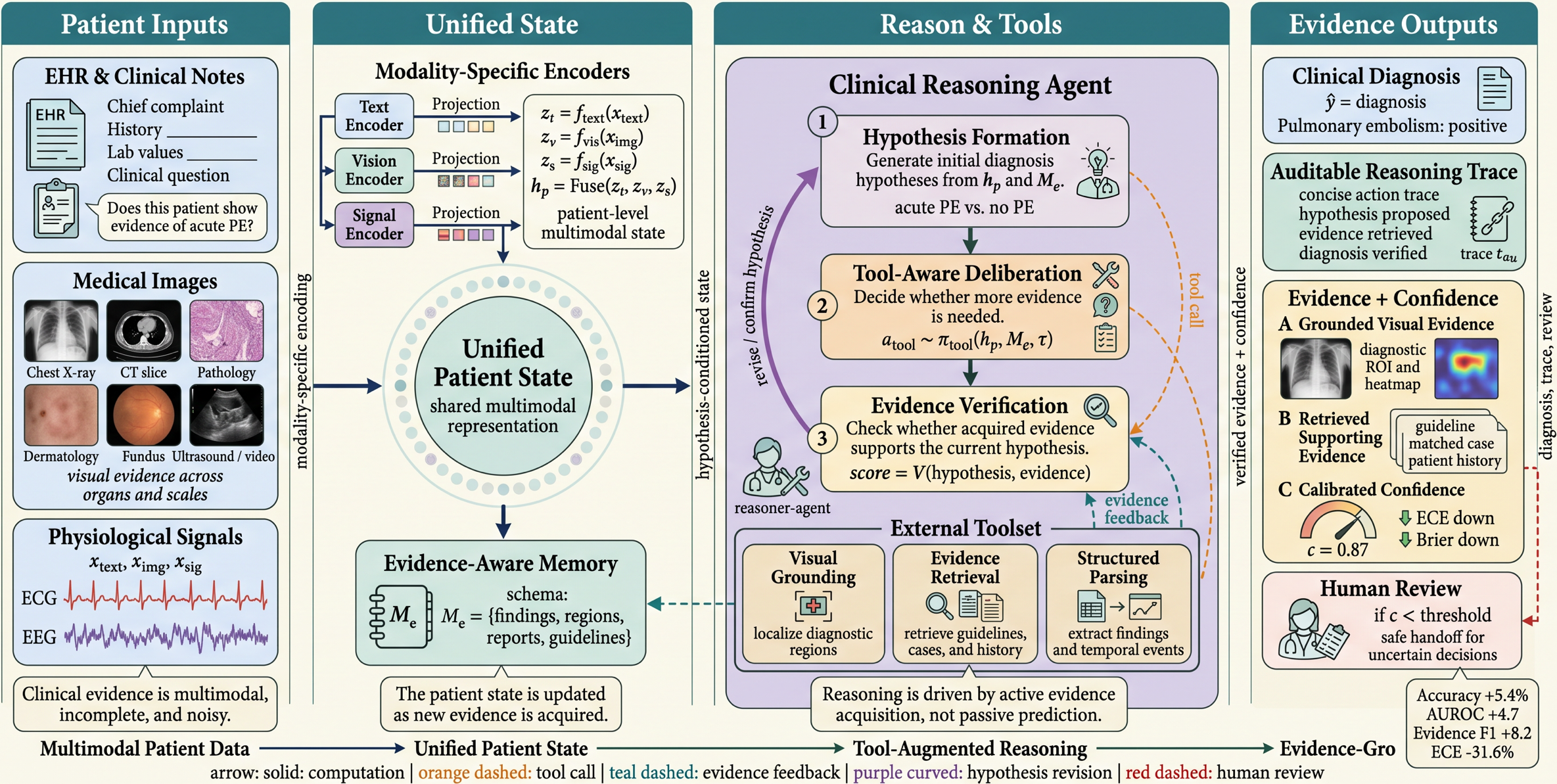}
\caption{Overview of MedTRACE. The framework integrates multimodal patient inputs into a unified patient state, performs hypothesis-driven clinical reasoning with external tools, verifies acquired evidence, and produces evidence-grounded diagnoses with auditable traces and calibrated confidence.}
\label{fig:medtrace}
\end{figure}

Figure~\ref{fig:medtrace} presents an overview of MedTRACE. The framework first encodes heterogeneous patient inputs into a unified state, then performs a closed loop of hypothesis formation, tool-aware deliberation, and evidence verification. Verified findings are accumulated in an evidence-aware memory and used to update the patient state and diagnostic hypothesis. The final output contains a diagnosis, grounded evidence, an auditable action trace, and a calibrated confidence score that determines whether human review is required.

\subsection{Problem Formulation and Patient State}

For each case, the input is $\mathcal{X}=\{x^{\mathrm{txt}},x^{\mathrm{img}},x^{\mathrm{sig}}\}$, containing clinical text, medical images, and physiological signals. The framework accesses visual grounding, retrieval, and structured parsing tools, denoted by $\mathcal{T}=\{\mathcal{T}_{\mathrm{vis}},\mathcal{T}_{\mathrm{ret}},\mathcal{T}_{\mathrm{parse}}\}$. Each available modality is encoded and projected into a shared space, and masked attention produces the initial patient state:
\begin{equation}
\begin{aligned}
z_m&=P_m(f_m(x^m)),\qquad
\alpha_m=\frac{b_m e^{q^\top\tanh(W_mz_m+e_m)}}
{\sum_j b_j e^{q^\top\tanh(W_jz_j+e_j)}},\\
h_p&=\operatorname{LN}\!\left(W_f\sum_m\alpha_mz_m\right),
\quad m\in\{\mathrm{txt},\mathrm{img},\mathrm{sig}\},
\end{aligned}
\label{eq:patient_state}
\end{equation}
where $b_m$ masks missing modalities. As evidence accumulates, the state is updated by attending to memory embeddings $E_t$:
\begin{equation}
h_t=\operatorname{LN}\!\left(
h_p+W_e\operatorname{Attn}(h_p,E_t,E_t)\right).
\label{eq:evidence_state}
\end{equation}
At step $t$, MedTRACE maintains $h_t$, diagnostic hypotheses $\mathcal{H}_t$, evidence memory $\mathcal{M}_t$, and trace $\tau_t$, and ultimately returns a diagnosis, supporting evidence, calibrated confidence, and a human-review flag.

\subsection{Hypothesis-Centered Tool-Aware Reasoning}

The agent predicts a differential diagnosis and an evidence-sufficiency score from the evidence-conditioned state:
\begin{equation}
\begin{aligned}
p_t(y)&=\operatorname{softmax}\!\left(
W_h[h_t;\operatorname{Pool}(\mathcal{M}_t)]+b_h\right),\\
s_t&=\sigma\!\left(
w_s^\top[h_t;\operatorname{Pool}(\mathcal{M}_t);
\operatorname{Ent}(p_t)]+b_s\right).
\end{aligned}
\label{eq:hyp_suff}
\end{equation}
The top-$K$ conditions form $\mathcal{H}_t$. If $s_t$ is low, the agent chooses visual grounding, retrieval, structured parsing, or stopping by balancing predicted utility, expected information gain, and tool cost:
\begin{equation}
a_t^*=\arg\max_{a\in\mathcal{A}}
\left[\log\pi_\theta(a\mid h_t,\mathcal{H}_t,\mathcal{M}_t)
+\beta\widehat{\operatorname{IG}}(a)-\lambda C(a)\right].
\label{eq:tool_policy}
\end{equation}
The selected tool localizes relevant image regions, retrieves guidelines or similar cases, or extracts structured findings and temporal events.

\subsection{Evidence Memory and Verification}

A tool returns $e_t=(v_t,\ell_t,s_t^e,m_t^e,q_t^e)$, representing its content, provenance, localization, type, and quality. The verifier classifies its relationship to each hypothesis as support, contradiction, or insufficient and stores relevant, non-duplicate evidence:
\begin{equation}
\begin{aligned}
v(e_t,h)&=\operatorname{softmax}\!\left(W_v\phi(e_t,h,h_t)\right),\\
\mathcal{M}_{t+1}&=\mathcal{M}_t\cup\{(e_t,r_t,\rho_t,v(e_t,h))\}.
\end{aligned}
\label{eq:verify_memory}
\end{equation}
Here, $\phi$ captures semantic relevance, source quality, cross-modal agreement, and localization overlap, while $\rho_t$ weights evidence reliability. Accumulated support $S_t(h)$ and contradiction $C_t(h)$ revise the diagnostic posterior:
\begin{equation}
\begin{aligned}
S_t(h)&=\sum_{e\in\mathcal{M}_t}\rho(e,h)v_{\mathrm{sup}}(e,h),\quad
C_t(h)=\sum_{e\in\mathcal{M}_t}\rho(e,h)v_{\mathrm{con}}(e,h),\\
p_{t+1}(h)&=
\frac{p_t(h)e^{\eta[S_t(h)-C_t(h)]}}
{\sum_{h'}p_t(h')e^{\eta[S_t(h')-C_t(h')]}}.
\end{aligned}
\label{eq:hyp_update}
\end{equation}
Contradictory evidence therefore lowers the current hypothesis posterior and triggers revision, implementing the feedback loop in Fig.~\ref{fig:medtrace}.

\subsection{Evidence-Grounded Decision and Confidence Calibration}

The agent stops when evidence is sufficient, the posterior is stable, and no unresolved contradiction remains. It selects the highest-probability diagnosis and retains only reliable supporting evidence. Confidence jointly reflects posterior probability, evidence coverage $Q_t$, support, and conflict:
\begin{equation}
\begin{aligned}
\hat{y}&=\arg\max_h p_t(h),\qquad
\mathcal{E}^{+}=\{e:v_{\mathrm{sup}}(e,\hat y)>\tau_v,\,
\rho(e,\hat y)>\tau_\rho\},\\
c&=\sigma\!\left(\frac{
w_c^\top[\operatorname{logit}p_t(\hat y);S_t(\hat y);Q_t;-C_t(\hat y)]+b_c}
{T_{\mathrm{cal}}}\right),\\
r_{\mathrm{human}}&=\mathbb{I}
[c<\tau_c\lor C_t(\hat y)>\tau_{\mathrm{con}}\lor t=T_{\max}].
\end{aligned}
\label{eq:decision}
\end{equation}
The trace $\tau$ records tool calls, evidence provenance, verification outcomes, and hypothesis revisions for audit.

\begin{algorithm}[t]
\caption{Evidence-grounded inference with MedTRACE}
\label{alg:medtrace}
\SetKwInput{KwIn}{Input}
\SetKwInput{KwOut}{Output}
\KwIn{Patient data $\mathcal{X}$, tools $\mathcal{T}$, budget $T_{\max}$}
\KwOut{$(\hat y,\mathcal{E}^{+},\tau,c,r_{\mathrm{human}})$}
$h_p\leftarrow\operatorname{EncodeFuse}(\mathcal{X})$;
$\mathcal{M}\leftarrow\varnothing$; $\tau\leftarrow\varnothing$\;
\While{$t<T_{\max}$}{
  $h_t\leftarrow\operatorname{Condition}(h_p,\mathcal{M})$\;
  $(\mathcal{H}_t,p_t,s_t)\leftarrow\operatorname{Reason}(h_t,\mathcal{M})$\;
  \If{$s_t\ge\tau_s$ and $\operatorname{Consistent}(\mathcal{H}_t,\mathcal{M})$}{
    \textbf{break}\;}
  $a_t\leftarrow\operatorname{SelectTool}(h_t,\mathcal{H}_t,\mathcal{M})$\;
  $e_t\leftarrow\mathcal{T}_{a_t}(\mathcal{X},\mathcal{H}_t,\mathcal{M})$\;
  $(v_t,\rho_t)\leftarrow\operatorname{Verify}(e_t,\mathcal{H}_t)$\;
  $\mathcal{M}\leftarrow\operatorname{Update}(\mathcal{M},e_t,v_t,\rho_t)$\;
  $p_{t+1}\leftarrow\operatorname{Revise}(p_t,\mathcal{M})$;
  $\tau\leftarrow\tau\cup\{(a_t,e_t,v_t)\}$\;
}
$\hat y\leftarrow\arg\max_h p_t(h)$;
$\mathcal{E}^{+}\leftarrow\operatorname{Support}(\hat y,\mathcal{M})$\;
$(c,r_{\mathrm{human}})\leftarrow\operatorname{CalibrateReview}
(\hat y,\mathcal{E}^{+},\mathcal{M})$\;
\Return{$(\hat y,\mathcal{E}^{+},\tau,c,r_{\mathrm{human}})$}\;
\end{algorithm}

\subsection{Training Objective}

MedTRACE jointly supervises the final diagnosis and intermediate evidence process:
\begin{equation}
\mathcal{L}=
\mathcal{L}_{\mathrm{diag}}
+\lambda_{\mathrm{tool}}\mathcal{L}_{\mathrm{tool}}
+\lambda_{\mathrm{ev}}\mathcal{L}_{\mathrm{ev}}
+\lambda_{\mathrm{ver}}\mathcal{L}_{\mathrm{ver}}
+\lambda_{\mathrm{gr}}\mathcal{L}_{\mathrm{gr}}
+\lambda_{\mathrm{cal}}\mathcal{L}_{\mathrm{cal}}.
\label{eq:objective}
\end{equation}
The terms denote cross-entropy losses for diagnosis and tool selection, binary evidence-selection loss, three-way consistency-verification loss, Dice--GIoU visual grounding loss, and Brier calibration loss, respectively. We first train the encoders and intermediate predictors with supervised targets and then optimize complete reasoning rollouts end to end.

\section{Experiments}

\subsection{Experimental Setup}

\noindent\ding{182}\enspace\textbf{Datasets and tasks.}
We evaluate three multimodal settings. For EHR--image diagnosis, we link MIMIC-IV with MIMIC-CXR at the patient level~\cite{johnson2023mimiciv,johnson2019mimiccxr}. For image--text reasoning, MIMIC-CXR supports diagnosis and textual evidence selection, while Chest ImaGenome and VinDr-CXR provide region annotations for visual grounding~\cite{wu2021chestimagenome,nguyen2022vindrcxr}. For signal--text reasoning, we pair 12-lead ECGs with reports or diagnostic labels from MIMIC-IV-ECG and PTB-XL~\cite{gow2023mimicivecg,wagner2020ptbxl}. We use official splits when available and otherwise adopt a patient-disjoint 70/10/20 split. Images are resized to $224\times224$, text is truncated to 512 tokens, and ECGs are standardized to 10 seconds at 500 Hz.

\smallskip
\noindent\ding{183}\enspace\textbf{Model and tools.}
We use BioClinicalBERT as the text encoder~\cite{alsentzer2019clinicalbert}, the ViT-B/16 branch of BiomedCLIP as the image encoder~\cite{zhang2024biomedclip}, and the ECG branch of CSFM as the signal encoder~\cite{gu2026cardiac}. All features are projected to 768 dimensions. The reasoning agent uses Qwen2.5-VL-7B-Instruct~\cite{bai2025qwen25vl} with LoRA ($r=16$, $\alpha=32$, dropout $=0.05$). BiomedParse provides visual grounding~\cite{zhao2025biomedparse}, MedCPT retrieves the top five guidelines, articles, or training-set cases~\cite{jin2023medcpt}, and BioClinicalBERT with scispaCy extracts structured findings~\cite{neumann2019scispacy}. Test cases and labels are excluded from the retrieval index.

\smallskip
\noindent\ding{184}\enspace\textbf{Optimization.}
Each case permits at most four reasoning rounds and three tool calls. We train for 20 epochs with AdamW, using learning rates of $2\times10^{-5}$ for pretrained components and $10^{-4}$ for newly initialized modules, batch size 16, weight decay 0.01, 5\% warm-up, and gradient clipping at 1.0. Early stopping uses validation Macro-F1 with patience 3. Experiments run on four NVIDIA A100 80GB GPUs with seeds 13, 42, and 2026; we report means and 95\% confidence intervals.

\subsection{Baselines and Evaluation Metrics}

\noindent\ding{182}\enspace\textbf{Baselines.}\par
\noindent We compare against single-modality specialists (BioClinicalBERT, DenseNet-121, and 1D-ResNet) and standard early/late fusion and cross-attention models. We also evaluate LLaVA-Med and MedGemma 1.5~\cite{li2023llavamed,sellergren2026medgemma}, with BiomedCLIP as a contrastive baseline~\cite{zhang2024biomedclip}. Tool-based baselines include RAG and ReAct~\cite{lewis2020rag,yao2023react}, together with MedAgent-Pro and CARE~\cite{wang2026medagent,du2026care}. A same-backbone ReAct baseline uses the identical Qwen2.5-VL model, tools, retrieval corpus, and call budget to isolate the contribution of evidence memory and consistency verification.

\smallskip
\noindent\ding{183}\enspace\textbf{Metrics.}
Diagnosis is evaluated with Accuracy, Macro-F1, and AUROC. Evidence selection uses Precision, Recall, and F1, while visual grounding uses IoU and Dice. Calibration is measured by ECE, Brier score, and NLL. We additionally report average tool calls, latency, cost, unsupported diagnostic error rate, and human-review rate. An unsupported error is an incorrect diagnosis for which the selected evidence does not match expert-annotated support. All principal results are averaged over three seeds.

\subsection{Main Results}

\begin{table}[t]
\centering
\caption{Aggregate diagnostic results across the three multimodal settings. Best and second-best results are shown in bold and underlined, respectively.}
\label{tab:main_results}
\small
\setlength{\tabcolsep}{9pt}
\rowcolors{2}{tablerow}{white}
\begin{tabular}{lccc}
\hline
\rowcolor{tableheader}
\textbf{Method} & \textbf{Accuracy (\%)} & \textbf{Macro-F1 (\%)} & \textbf{AUROC} \\
\hline
BioClinicalBERT & 68.9 & 66.8 & 0.772 \\
DenseNet-121 & 70.4 & 68.7 & 0.786 \\
Early Fusion & 72.6 & 70.9 & 0.801 \\
Cross-Attention & 74.8 & 72.7 & 0.816 \\
LLaVA-Med & 74.1 & 71.9 & 0.811 \\
MedGemma 1.5 & 76.3 & 73.8 & 0.826 \\
RAG Agent & 75.7 & 73.1 & 0.823 \\
ReAct Agent & 76.8 & 74.5 & 0.831 \\
MedAgent-Pro & 77.1 & 74.9 & 0.834 \\
CARE & \underline{77.5} & \underline{75.2} & \underline{0.837} \\
\hline
\textbf{MedTRACE} & \textbf{81.7} & \textbf{79.8} & \textbf{0.884} \\
\hline
\end{tabular}
\end{table}

\begin{figure}[t]
\centering
\includegraphics[width=\textwidth]{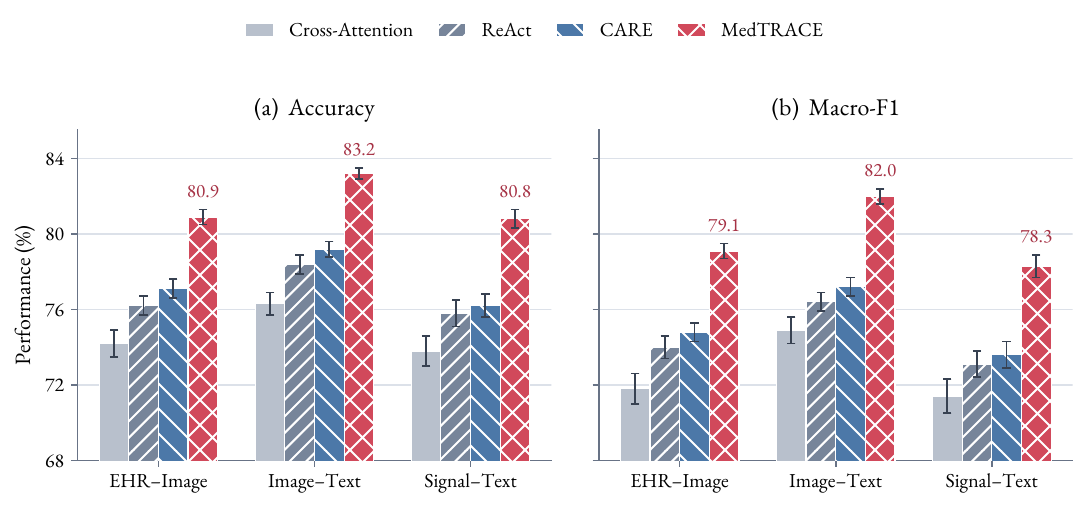}
\caption{Cross-task diagnostic performance of representative fusion, reasoning-agent, and clinical-agent baselines. Error bars denote standard deviations over three seeds.}
\label{fig:cross_task_results}
\end{figure}

Table~\ref{tab:main_results} reports the aggregate results. MedTRACE achieves the best performance on all diagnostic metrics, reaching 81.7\% Accuracy, 79.8\% Macro-F1, and 0.884 AUROC. Relative to the strongest baseline, CARE, it improves Accuracy by 5.4\% and AUROC by 4.7 percentage points. Figure~\ref{fig:cross_task_results} further shows consistent gains across modality combinations: MedTRACE improves Accuracy over CARE by 3.8--4.6 points and Macro-F1 by 4.3--4.8 points. The largest Accuracy gain occurs in signal--text reasoning, where explicit verification is especially useful for integrating heterogeneous temporal evidence. Its advantage over the same-backbone ReAct Agent under an identical toolset, retrieval corpus, and call budget indicates that the overall gains arise from hypothesis-centered tool selection, evidence memory, and consistency verification rather than additional tools or a stronger backbone.

\subsection{Evidence Quality, Calibration, and Safety}

\begin{table}[t]
\centering
\caption{Evidence quality, calibration, and safety results. UDE denotes the unsupported diagnostic error rate. Higher is better for Evidence F1, IoU, and Dice; lower is better for the remaining metrics.}
\label{tab:evidence_safety}
\scriptsize
\resizebox{\textwidth}{!}{%
\rowcolors{2}{tablerow}{white}
\begin{tabular}{lcccccccc}
\hline
\rowcolor{tableheader}
\textbf{Method} & \textbf{Evid. F1} & \textbf{IoU} & \textbf{Dice} &
\textbf{ECE} & \textbf{Brier} & \textbf{NLL} &
\textbf{UDE (\%)} & \textbf{Review (\%)} \\
\hline
ReAct & 65.9 & 56.2 & 63.8 & 0.121 & 0.165 & 0.489 & 15.8 & 18.7 \\
MedAgent-Pro & 68.1 & 59.4 & 66.5 & 0.108 & 0.154 & 0.462 & 13.4 & 16.5 \\
CARE & \underline{70.4} & \underline{61.8} & \underline{68.7} &
\underline{0.095} & \underline{0.142} & \underline{0.426} &
\underline{11.5} & \underline{14.8} \\
\hline
\textbf{MedTRACE} & \textbf{78.6} & \textbf{68.3} & \textbf{74.9} &
\textbf{0.065} & \textbf{0.116} & \textbf{0.351} &
\textbf{8.3} & \textbf{12.6} \\
\hline
\end{tabular}
}
\end{table}

\begin{figure}[t]
\centering
\includegraphics[width=\textwidth]{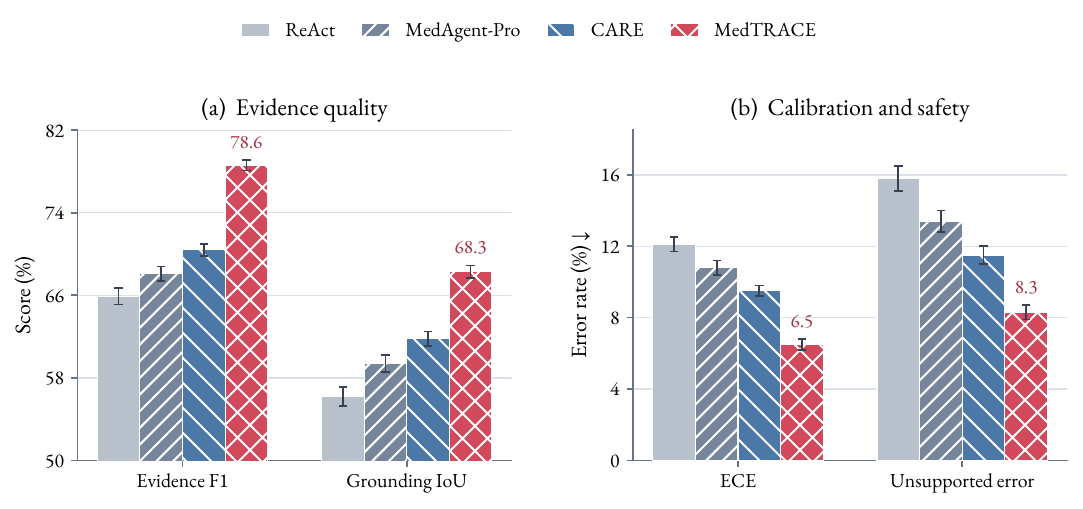}
\caption{Comparison of evidence quality, confidence calibration, and diagnostic safety. Error bars denote standard deviations over three seeds.}
\label{fig:evidence_safety}
\end{figure}

Table~\ref{tab:evidence_safety} and Fig.~\ref{fig:evidence_safety} show that the diagnostic gains are accompanied by stronger evidence grounding and reliability. Relative to CARE, MedTRACE improves evidence-selection F1 by 8.2 points and visual-grounding IoU by 6.5 points, indicating more complete and spatially precise support for its predictions. Its ECE decreases from 0.095 to 0.065, a 31.6\% relative reduction, with consistent improvements in Brier score and NLL. MedTRACE also reduces unsupported diagnostic errors from 11.5\% to 8.3\% (27.8\%) while lowering the human-review rate from 14.8\% to 12.6\%. This joint reduction suggests that calibrated evidence verification avoids both unsafe autonomous decisions and unnecessary escalation.

\subsection{Ablation Study}

\begin{table}[t]
\centering
\caption{Ablation results averaged across the three multimodal settings. Higher is better for all metrics except ECE.}
\label{tab:ablation}
\scriptsize
\resizebox{\textwidth}{!}{%
\rowcolors{2}{tablerow}{white}
\begin{tabular}{lccccc}
\hline
\rowcolor{tableheader}
\textbf{Variant} & \textbf{Acc. (\%)} & \textbf{Macro-F1 (\%)} &
\textbf{AUROC} & \textbf{Evidence F1 (\%)} & \textbf{ECE} $\downarrow$ \\
\hline
w/o Tool Selection & 79.4 & 77.3 & 0.858 & 73.1 & 0.079 \\
w/o Evidence Memory & 78.8 & 76.6 & 0.851 & 71.8 & 0.083 \\
w/o Consistency Verification & 78.2 & 75.9 & 0.846 & 70.9 & 0.087 \\
w/o Confidence Calibration & 81.5 & 79.6 & 0.882 & 78.4 & 0.101 \\
\hline
\textbf{MedTRACE} & \textbf{81.7} & \textbf{79.8} & \textbf{0.884} &
\textbf{78.6} & \textbf{0.065} \\
\hline
\end{tabular}
}
\end{table}

Table~\ref{tab:ablation} demonstrates that the components of MedTRACE make complementary contributions. Removing consistency verification causes the largest diagnostic degradation, reducing Accuracy by 3.5 points, which confirms the importance of explicitly distinguishing supporting, contradictory, and insufficient evidence when revising hypotheses. Removing evidence memory decreases evidence-selection F1 by 6.8 points, showing that structured cross-modal evidence accumulation improves coverage and reduces fragmented reasoning. Tool-aware selection also provides substantial gains over a fixed tool policy. Confidence calibration has little effect on discrimination but is essential for reliability: without it, ECE increases from 0.065 to 0.101. These results jointly support the complete closed-loop design.

\subsection{Efficiency Analysis}

\vspace{-5pt}
\begin{table}[H]
\centering
\caption{Average inference efficiency per case. Cost includes measured GPU time and external-tool execution. Lower is better for calls, latency, and cost.}
\label{tab:efficiency}
\small
\setlength{\tabcolsep}{8pt}
\rowcolors{2}{tablerow}{white}
\begin{tabular}{lcccc}
\hline
\rowcolor{tableheader}
\textbf{Method} & \textbf{Tool Calls} & \textbf{Latency (s)} &
\textbf{Cost (\$)} & \textbf{Accuracy (\%)} \\
\hline
RAG Agent & \textbf{2.0} & \textbf{2.9} & \textbf{0.014} & 75.7 \\
ReAct Agent & 3.0 & 4.8 & 0.024 & 76.8 \\
MedAgent-Pro & 3.4 & 5.5 & 0.029 & 77.1 \\
CARE & 3.2 & 5.1 & 0.027 & 77.5 \\
\hline
MedTRACE & 2.4 & 4.1 & 0.021 & \textbf{81.7} \\
\hline
\end{tabular}
\end{table}

\vspace{-14pt}
\noindent
\begin{minipage}[t]{0.43\textwidth}
\vspace{0pt}
Table~\ref{tab:efficiency} and Fig.~\ref{fig:efficiency} show that MedTRACE offers the best accuracy--efficiency trade-off. Compared with CARE, it reduces tool calls by 25.0\%, latency by 19.6\%, and cost by 22.2\%, while improving Accuracy by 4.2 points. It also exceeds the same-backbone ReAct Agent by 4.9 points with fewer calls. Although RAG is cheaper, its accuracy is substantially lower; MedTRACE therefore lies on the Pareto frontier.
\end{minipage}
\hfill
\begin{minipage}[t]{0.54\textwidth}
\vspace{0pt}
\centering
\includegraphics[width=\linewidth]{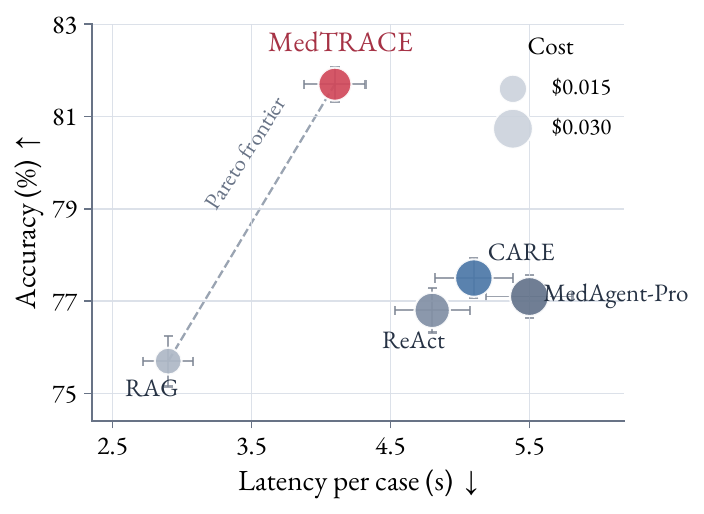}
\vspace{-5pt}
\refstepcounter{figure}
\label{fig:efficiency}
{\small\textbf{Fig.~\thefigure.} Accuracy--efficiency trade-off. Bubble area denotes per-case cost; error bars show standard deviations over three seeds.}
\end{minipage}

\smallskip
\noindent The efficiency advantage follows from evidence-aware stopping: straightforward cases terminate once high-quality support is sufficient, whereas ambiguous cases retain the remaining tool budget. This adaptive allocation avoids the uniform cost of fixed tool pipelines and concentrates computation where additional evidence can change the diagnosis.

\subsection{Qualitative Case Study}

\begin{figure}[t]
\centering
\includegraphics[width=\textwidth]{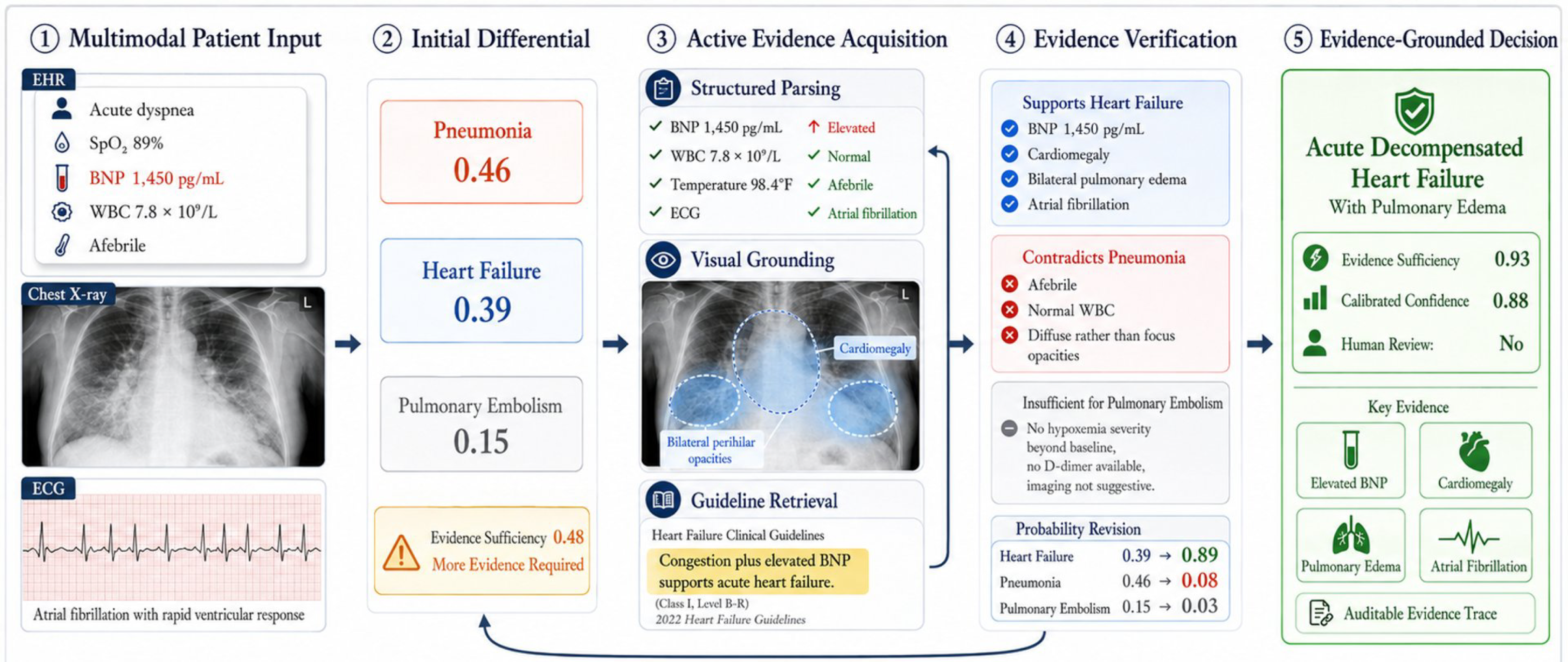}
\caption{Qualitative case study of evidence-grounded hypothesis revision. MedTRACE integrates EHR, chest X-ray, and ECG findings, acquires targeted evidence with external tools, and revises an initial pneumonia hypothesis to acute decompensated heart failure with pulmonary edema.}
\label{fig:case_study}
\end{figure}

Figure~\ref{fig:case_study} illustrates a representative multimodal case. Pulmonary opacities initially favor pneumonia, but an evidence-sufficiency score of 0.48 triggers further investigation. Structured parsing identifies elevated BNP, normal WBC, absence of fever, and atrial fibrillation; visual grounding localizes cardiomegaly and bilateral perihilar opacities; and guideline retrieval supports acute heart failure. Verification therefore raises the heart-failure posterior from 0.39 to 0.89 and yields calibrated confidence of 0.88 without human review, demonstrating auditable revision of an early hypothesis.

\section{Conclusion}

We present MedTRACE, a tool-augmented multimodal clinical reasoning agent that formulates diagnosis as an iterative process of hypothesis formation, active evidence acquisition, and consistency verification. By integrating modality-specific patient representations with tool-aware deliberation and evidence memory, MedTRACE produces diagnoses supported by traceable evidence and calibrated confidence. Experiments across EHR--image, image--text, and signal--text settings show consistent improvements in diagnostic accuracy, evidence selection, visual grounding, and calibration, while reducing unsupported diagnostic errors and unnecessary tool use. Ablation results further demonstrate the complementary roles of tool selection, evidence memory, consistency verification, and confidence calibration. These findings indicate that explicitly reasoning about evidence sufficiency is an effective path toward more reliable and auditable multimodal clinical decision support. Future work will evaluate MedTRACE prospectively across institutions while accounting for privacy--utility trade-offs in data sharing~\cite{duan2026gaitprotector}. We will also extend its tool library and modality coverage and study safe adaptation under distribution shift and evolving clinical guidelines.

%
%
\bibliographystyle{splncs04}
\bibliography{ref}

\end{document}